# Probabilistic Inference and Non-Monotonic Inference*

Henry E. Kyburg, Jr.
University of Rochester
Rochester N.Y. 14620

## 1. Introduction.

(1) I have enough evidence to render the sentence $S$ probable.

(2) So, relative to what I know, it is rational of me to believe $S$.

(3) Now that I have more evidence, $S$ may no longer be probable.

(4) So now, relative to what I know, it is not rational of me to believe $S$.

These seem a perfectly ordinary, common sense, pair of situations. Generally and vaguely, I take them to embody what I shall call *probabilistic inference*. This form of inference is clearly non-monotonic. Very few people in AI seem to have looked at it carefully.

There are exceptions: Jane Nutter (1987) thinks that sometimes probability has something to do with non-monotonic reasoning. Judea Pearl has recently (1987) been exploring the possibility.

There are any number of people whom one might call probability enthusiasts. Cheeseman (1985), Henrion (1987) and others think it useful to look at a distribution of probabilities over a whole algebra of statements, to update that distribution in the light of new evidence, and to use the latest updated distribution of probability over the algebra as a basis for planning and decision making. A slightly weaker form of this approach is captured by Nilsson (1986), where one assumes certain probabilities for certain statements, and infers the probabilities, or constraints on the probabilities, of other statements.

None of this corresponds to what I call probabilistic inference. All of the inference that is taking place is strictly <u>deductive</u>. Deductive inference, particularly that concerned with the distribution of classical probabilities or chances, is of great importance. But this is not to say that there is not also an important role for what logicians have called "ampliative" or "inductive" or "scientific" inference, in which the conclusion goes <u>beyond</u> the premises, asserts more than do the premises. This depends on what David Israel (1980) has called "real rules of inference." It is characteristic of any such logic or inference procedure that it can go

229

wrong: that statements accepted at one point may be rejected at a later point.

## 2. McCarthy and Hayes.

As a matter of historical conjecture, I would suggest that it is the enormously influential article by John McCarthy and Pat Hayes (1969) that slowed the exploration of probabilistic inference. They offer powerful arguments against the use of probability as an approach to non-monotonicity. Since much of this argument applies equally well to most formalizations of non-monotonic inference, it is worth rehearsing their objections.

They commence with first order predicate calculus, and add 3 operators: Consistent($\emptyset$), Normally($\emptyset$), Probably($\emptyset$). They consider a set $\sigma$ of sentences, and add new sentences to it according to the rules:
1. Any consequence of $\sigma$ may be added [to $\sigma$]
2. If $\emptyset$ is consistent with $\sigma$, consistent($\emptyset$) may be added.
3. Normally($\emptyset$),Consistent($\emptyset$)⊢ Probably($\emptyset$)
4. $\emptyset$ ⊢ Probably($\emptyset$) is a possible deduction
5. If $\emptyset_1,...,\emptyset_n$ ⊢ $\emptyset$ is a possible deduction, then Probably($\emptyset_1$),..., Probably($\emptyset_n$) ⊢ Probably($\emptyset$) is a possible deduction.

Two objections are offered against the use of probabilities:

"It is not clear how to attach probabilities to statements containing quantifiers in a way that corresponds to the amount of conviction people have." (p. 490)

"The information necessary to assign numerical probabilities is not ordinarily available.... Therefore ... epistemologically inadequate." (p.490)

A propos of these objections we only remark that Gaifman (1964), Scott and Krauss (1966), and others have provided schemes for assigning probabilities to quantified statements; and if we don't demand exactitude, we may have enough information to assign (approximate) probabilities.

The example that McCarthy and Hayes offer is more telling:
$P$ looks up $Q$'s number;
So he knows it;
So he should believe that if he dials it, "he will come into
conversation with $Q$."

In order to obtain this conclusion in terms of probability, we need to add to the rules:
6.     Probably($\emptyset$) ⊢ $\emptyset$

We also need a rule for deleting statements from $\sigma$. But the details of these rules are not to the point. It is rather the intuitions that lie behind them (if not the axioms themselves) that have guided work in non-monotonic inference.

Rules (1) and (5) constitute the main stumbling blocks in the way of developing a system to embody these intuitions, even without (6). In particular, rule (1) requires that $\sigma$ be strictly consistent; we shall argue later that this is neither necessary nor desirable.

From (1), (3), (4) and (5) and natural assumptions we get a lottery paradox (Kyburg 1961). (With (6) it becomes a contradiction.)

Assume: $\emptyset_i$: ticket $i$ loses; $\emptyset$: all tickets lose; $\sigma$ contains "$\sim\emptyset$"; Normally($\emptyset_i$); Consistent($\emptyset_i$);

By (3), Probably($\emptyset_i$). By (1) $\{\emptyset_i\} \vdash \emptyset$. Therefore by (5), Probably($\emptyset$). But by (4), Probably($\sim\emptyset$). If this isn't contradictory enough, add (6). Note that nothing self-referential is involved.

### 3. Non-monotonic Inference.

In this section we will show that three well-known examples of efforts to codify non-monotonic inference stumble equally over the lottery paradox. Before doing so, it will be well to consider the question of what we want of this set $\sigma$ of "believed" statements to do for us. Statements on that list should serve as <u>evidence</u> for other statements. They should be useful as premises in <u>planning</u> (the agent $P$ is constructing a reasonable plan for talking to $Q$). They should define the limits (at a given point in the collection of evidence) of <u>reasonable</u> or <u>serious possibility</u>. It is the planning and designing function of $\sigma$ that is of clearest importance in AI. Don Perlis (1980) has come up with the term "use-beliefs" to describe this set of sentences. Perlis has also argued that standard systems of non-monotonic reasoning have difficulty in handling iterations of application. His zookeeper example -- one of the birds is sick and can't fly -- is just the lottery paradox made realistic. But Perlis still seems to want $\sigma$ to be consistent.

Let us look at the system of Reiter (1980), because it embodies the intuitions driving default logic particularly clearly.

The intuitive idea is to adopt a set of defaults **D** inducing an extension $E$ of some underlying incomplete set of wffs $W$. The extension $E$ need not be unique, but (1), (2) and (3) should be satisfied:

(1) $W \vdash E$,
(2) $E$ is deductively closed, and
(3) Suppose $(A:MB_1,...MB_n/C)$ is a default, and each

231

of $B_j$ is consistent with $E$. Then $C$ belongs to $E$.

The lottery with $n$ tickets: $W$ = {sentences describing a fair lottery}; a perfectly natural default is :$M^\sim \emptyset_j/\emptyset_j$ -- if you don't know that ticket $i$ wins, i.e., it's reasonable to believe it loses.

This gives you $n$ extensions $E_j$, each of which specifies one winner and $n$-1 losers. That seems implausible; it is clearly not useful for planning to have a number of extensions. Note that it is deductive closure that prevents our having a single extension containing each $\emptyset_j$. Deductive closure would lead to $E$ containing all sentences.

How does circumscription handle the lottery tickets? Roughly speaking: If $x$ is a ticket, and $x$ is not abnormal, then $x$ loses the lottery. It is hard to know <u>exactly</u> what one can conclude, since (McCarthy 1980) we "clearly" have to include domain dependent heuristics for deciding what circumscriptions to make and when to take them back. But waiving difficulties of quantification, it certainly seems right that in the case of just one ticket, we should be able to conclude that it does not win the lottery.

The question then is, "How often can we iterate this argument?" It would seem that the answer is roughly n/2: If we have considered less than half the tickets, the next ticket is still (statistically) likely to lose, relative to what we have taken ourselves to know.

The non-monotonic logic of McDermott and Doyle (1980) construes the set of non-monotonic theorems as (roughly) the smallest fixed point under non-monotonic derivability. So it would seem that in lottery cases there would be no useful set of non-monotonic conclusions. And yet the lottery should not be dismissed as frivolous: consider the lottery as standing for any situation in which a certain outcome is taken to be "incredible", and consider an arbitrarily long sequence of such situations. We shall consider a specific case shortly.

### 4. The Cannonical Examples.

Before considering the seriousness of the inability of standard non-monotonic inference systems to deal with the lottery, let us show that probabilities do work reasonably well on the standard examples of non-monotonic argument.

We assume that probabilities depend on our knowledge of frequencies or chances, and further than they depend on what we know about the object at issue.

Let $E$ be evidence, $K$ be the set of acceptable beliefs.

(1) Tweety.

> We know "almost all birds fly" in $E$.
> We know "all penguins are birds" in $E$.
> We know "no penguins fly" in $E$.

Suppose we add to $E$ "Tweety is a bird." The probability that Tweety flies is high. High enough for us to accept "Tweety flies" in our body of knowledge (or use-beliefs) $K$.

Suppose we now add: "Tweety is a penguin" to $E$. The probability that Tweety flies, relative to $E$ is now low. We must <u>delete</u> "Tweety flies." And add "Tweety doesn't fly."

What we are doing here is using the <u>more specific reference class</u> as the basis for our probability. This principle is also given by McCarthy (1986), Etherington (1987) (as 'inferential distance'), Poole (1985) ("strictly more specific), and others.

(2) Nixon Diamond:

Add "Nixon is a Quaker" to the standard $E$. It is then probable relative to $E$ that Nixon is a pacifist; we add "Nixon is a Pacifist" to $K$ -- "we plan on it."

Now add "Nixon is a Republican" to the standard $E$. Similar: we add "Nixon is not a pacifist" to $K$.

Now add both. Knowledge about republicans and knowledge about quakers now conflict as a basis for asserting (or not) that Nixon is a pacifist. And we just don't know about the intersection. So we can't conclude anything. This conforms to the usual non-monotonic treatment. McCarthy (1986, p. 90) seems to give us a little more: we can conclude that Nixon is either an abnormal quaker or an abnormal republican. But the cash value of that is just that we don't know whether or not he is a pacifist.

(3) Cohabitation: (Reiter 1980)

$\text{Spouse}(x,y) \& \text{hometown}(y)=z : M \text{ Hometown}(x) = z / \text{hometown}(x) = z$
$\text{Employer}(x,y) \& \text{Location}(y)=z : M \text{ hometown}(x) = z / \text{hometown}(x) = z$

Consider John, whose spouse lives in Toronto and whose employer is located in Vancouver. We can derive an extension in which his hometown is Toronto, and one in which his hometown is in Vancouver.

Probability lets us conclude the disjunction: John's hometown is either Toronto or Vancouver. This may be false, of course, but that is the nature of probabilistic inference. The example suggests to Reiter: (1) ordering defaults (on what basis?) and (2) getting more information (a copout).

Reiter (1980) also suggests that default logic is useful for dealing with the frame problem. We use a default that says everything stays same

233

unless it is deducible that it has changed. Probability does even better here: When we add a new statement to $E$, it will not change most of the probabilities from which $K$ is derived. So without a default: everything in $k$ stays there unless it's <u>probability</u> has changed.

This brings up the most serious charge against probable inference. Since what is probable depends on our whole body of evidence $E$, any addition to $E$ requires the recomputation of all the probabilities on which $k$ is based. But non-monotonic approaches require the same thing: "networks must be reconditioned after each update" (Touretzky (1986); Etherington (1987).) How expensive that reconditioning is depends on how difficult it is to compute probabilities.

## 5. Consistency.

How can $K$ (the σ of McCarthy and Hayes) serve as a standard of serious possibility, as a basis for planning or designing, if it is inconsistent? Clearly, if it is deductively closed, it cannot. But it membership in $K$ is determined by high probability it <u>won't</u> be deductively closed. On the natural assumption that statements known to have the same truth value will have the same probability, it will contain (in principle) the logical consequences of statements that it contains. (Since if $P$ entails $Q$, $P \vdash P \& Q$ will be logically true and thus part of the evidence $E$; and the probability of $Q$ must be at least as great as that of $P \& Q$.)

Inconsistency is a good reason for avoiding deductive closure. But what is a good reason for having an inconsistent $K$? The lottery is frivolous. Is there a practical and natural counterpart? The answer is yes, and it comes from the eminently respectable scientific domain of measurement. Suppose my job is conducting scientific measurements of a certain sort -- blood sugar, say. It is clear that however sophisticated my measurements, it cannot be demanded that they be error free. Furthermore, there is no way of bounding measurement error. What <u>can</u> be demanded is that of each measurement, the probability that it is in error by more than .05 units be less than (say) .001. For purposes of further and for purposes of planning, it is obvious that I must believe, of each measurement, that it is within the permissible error of .05.

It is also clear that I ought also to believe -- accept as a serious possibility, accept for planning or design -- that of a large number ($10^6$?) measurements, at least one will not be within its permissible error. Thus my beliefs -- my use-beliefs -- are flatly inconsistent. The deductive closure of $K$ contains all the statements in the language. Yet I have no difficulty using this (unclosed!) set of statements $K$ for planning, or as a

234

standard of serious possibility. Even though the <u>conjunction</u> of statements in $K$ is impossible, it serves as a standard of serious possibility in the sense that if a statement contradicts a <u>member</u> of $K$, it is not to be regarded as a serious possibility. That a particular one of the measurements is not OK is not a <u>serious</u> possibility, though of course it is a possibility. And it is not a serious possibilty that <u>all</u> the measurements are within tolerance.

The probabilistic version of non-monotonic inference faces one irrefutable complaint. Real probability enthusiasts will say that the story I have just told, while reflecting the way in which people do talk, should be regarded simply as a rough approximation to the real truth which can only be represented by a probability distribution over all the states of the world (or all the sentences of the language). In particular, in planning to base some action on the value of my measurement, what I must "really" be doing is evaluating the probability that it is within tolerance, multiplying that by the utility of performing that action on that assumption, <u>and adding to that the negative expectation, similarly computed, of performing that action on the assumption that the measurement is not within tolerance.</u> If the probability of error .25, and the number of measurements involved were 10, clearly the probabilistic analysis would be preferable. Why not when the probabilities are smaller and the numbers bigger?

To this one can reply by waving one's hands at the computations involved: acceptance into the body of beliefs $K$ is not only realistic (as a representation of what humans do) but is the only computationally feasible way to go about these things. Maybe. Here it seems to me that we will only be able to get good answers about which is the best way to represent practical knowledge or use-beliefs by constructing systems that embody each approach.

## 6. <u>Conclusions.</u>

We have not provided an algorithm for performing probabilistic inference, though some efforts have been made along these lines (Loui 1986). We have not argued that pure probabilism is unworkable in principle. We have argued that many of the usual arguments against probabilistic inference are equally applicable to other forms of non-monotonic reasoning. We have argued that, in fact, once the single hurdle of "inconsistency" is overcome, probabilistic inference offers advantages (in some contexts) or at least no disadvantages (in most contexts) compared to non-monotonic reasoning. We have argued that, in fact, the ability to live comfortably with certain sorts of inconsistency is an important feature of probabilistic inference, and that it allows us to take as a basis for planning exactly those statements that are practically certain.

All this is not to say that the various forms of non-monotonic reasoning that have been explored are not useful for special purposes. It does suggest both that specific instances of non-monotonic argument can be justified by reference to probabilities, and that any sort of inference that was incompatible with probabilistic inference would have some serious strikes against it.

\* Research underlying this paper has been partially supported by the Signals Warfare Center of the U. S. Army.